\documentclass[a4paper, oneside]{article}
\pdfoutput=1

\usepackage{filecontents}

\usepackage{float}
\usepackage{multirow}
\usepackage[labelfont=bf]{caption}
\usepackage[utf8]{inputenc}
\usepackage[hidelinks]{hyperref}
\usepackage[a4paper,margin=1.25in]{geometry}
\usepackage{textcomp}
\usepackage{graphicx}
\usepackage{tabularx}
\usepackage{verbatim}
\usepackage{amsmath}
\usepackage{transparent} %used in header
\usepackage[svgnames]{xcolor}
\usepackage{sectsty}
\sectionfont{\color[rgb]{0.95,0.4,0.4}} %Tomato  % sets colour of chapters
\subsectionfont{\color[rgb]{0.95,0.4,0.4}}  % sets colour of sections
\chapterfont{\color[rgb]{0.95,0.4,0.4}}  
\usepackage{amssymb}

\usepackage{titlesec}
\titleformat{\chapter}{\normalfont\huge \color[rgb]{0.95,0.4,0.4}}{\thechapter.}{20pt}{\huge\bf}

\usepackage{fancyhdr}
\fancypagestyle{illumr}{
	\fancyhf{}
	\pagestyle{fancy}
	\setlength{\headheight}{15.2pt}
	\fancyhead[L]{\transparent{0.5}\includegraphics[height=0.6cm]{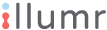}}
	\fancyhead[R]{\color{black!50} \thepage}
	
	\fancyfoot[L]{\color{black!50} Copyright \textcopyright \vspace{6pt} 2019 illumr ltd}
}

\fancypagestyle{plain}{%
	\fancyhf{}
	\pagestyle{fancy}
	\setlength{\headheight}{15.2pt}
	\fancyhead[L]{\transparent{0.5}\includegraphics[height=0.6cm]{logo.png}}
	\fancyfoot[L]{\color{black!50} Copyright \textcopyright \vspace{6pt} 2019 illumr ltd}
}



\let\stdsection\section
\renewcommand\section{\clearpage\stdsection}

\title{The relationship between Biological and Artificial Intelligence}

\author{ George \v{C}evora PhD \\ Chief Data Scientist, illumr Ltd.\\george.cevora@illumr.com}  
\date{}

\usepackage{tcolorbox}
\usepackage{wrapfig}
\usepackage{eurosym}
\begin{document}

\maketitle
\begin{abstract}
	Intelligence can be defined as a predominantly human ability to accomplish tasks that are generally hard for computers and animals. Artificial Intelligence [AI] is a field attempting to accomplish such tasks with computers. AI is becoming increasingly widespread, as are claims of its relationship with Biological Intelligence.  Often these claims are made to imply higher chances of a given technology succeeding, working on the assumption that AI systems which mimic the mechanisms of Biological Intelligence should be more successful.

In this article I will discuss the similarities and differences between AI and the extent of our knowledge about the mechanisms of intelligence in biology, especially within humans. I will also explore the validity of the assumption that biomimicry in AI systems aids their advancement, and I will argue that existing similarity to biological systems in the way Artificial Neural Networks [ANNs] tackle tasks is due to design decisions, rather than inherent similarity of underlying mechanisms. This article is aimed at people who understand the basics of AI (especially ANNs), and would like to be better able to evaluate the often wild claims about the value of biomimicry in AI.
\end{abstract}

\renewcommand{\abstractname}{Acknowledgements}
\begin{abstract}
I thank Kate Wilkinson for extensive editing  and proofreading of this article. Dr Kristjan Kalm's critical review of this article was essential for its scientific integrity; however,  not all views expressed in this article reflect Kristjan's own views. Finally, I would like to thank illumr Ltd, especially its founder and CEO Jason Lee, for sponsoring this work.
\end{abstract}

\section{From Symbolic AI to Machine Learning}
Symbolic AI was the prevailing approach to AI until the early 90’s. It is reliant on human programmers coding complex rules to enable machines to complete complex tasks. Continuing failure of this approach to solve many tasks crucial to intelligence provides a good contrast with Machine Learning – an alternative approach to AI which is essential to the current advent of artificially intelligent machines. 

In 1994 the reigning chess champion Garry Kasparov was beaten by Deep Blue. This was a great success for IBM at the time, and also for Artificial Intelligence. However, Deep Blue, while capable of playing world-class chess, could not move the pieces – a human helper was required for this task. We will start the exploration of the relationship between Artificial and Biological Intelligence by investigating the distinction between the tasks of deciding about chess moves and physically moving the pieces.

While chess is generally considered a difficult game, its rules are actually very simple and can easily fit on a single A4.  Deep Blue was programmed to calculate a lot of possible moves ahead according to those rules. Out of those moves Deep Blue just picked the most beneficial one. The reason why it took until 1994 for a computer to beat a human chess champion was a limitation on computing power. While the rules of chess are simple, there are very many possible moves - exploring even a small fraction of these requires a gargantuan amount of computation\footnote{The success of Deep Blue was actually for a significant part also due to making decisions which lines of game to avoid exploring. This fact doesn't have any bearing on the present argument though.}. Computers were not able to make this many calculations within a realistic time-frame before Deep Blue, but these days the computational power of Deep Blue is roughly matched by an average smartphone. Because the rules are so simple anyone with elementary coding skills can write a chess program and given sufficient computational power can beat world chess champion simply by applying the rules. This approach is called symbolic AI because the computer arrives to a solution by following predefined rules (commonly called symbolic computation).

\begin{figure}[p]
	\centering
	\begin{tabular}{cc}
		\includegraphics[width=7cm]{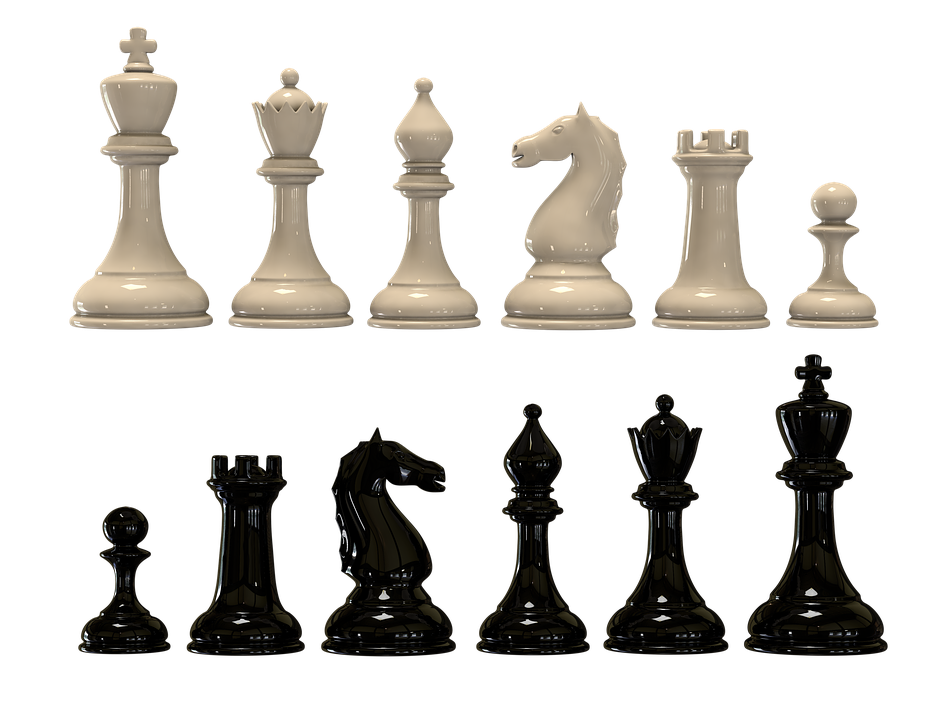} & \includegraphics[width=7cm]{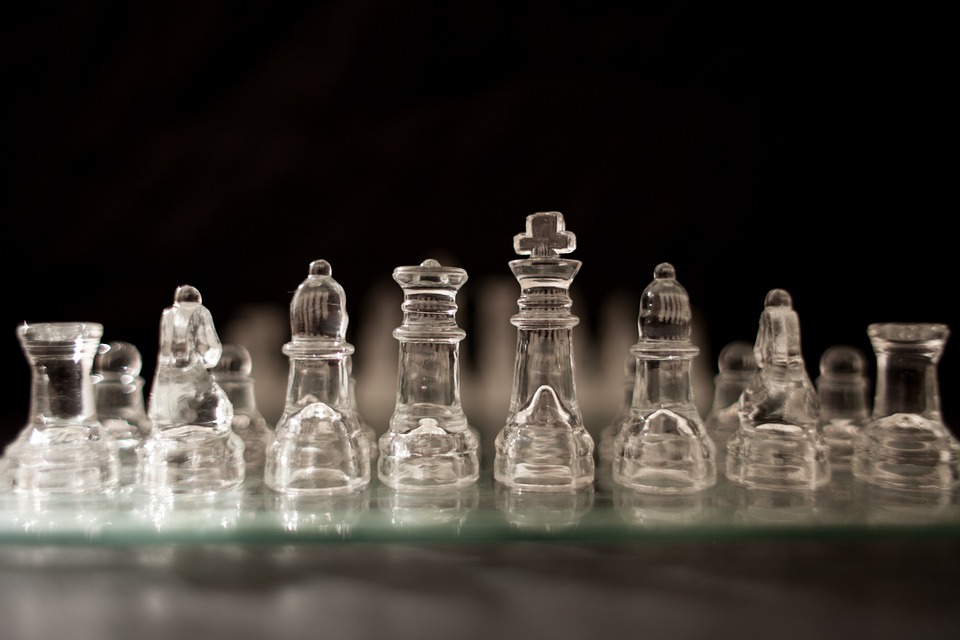}\\
		\includegraphics[width=7cm]{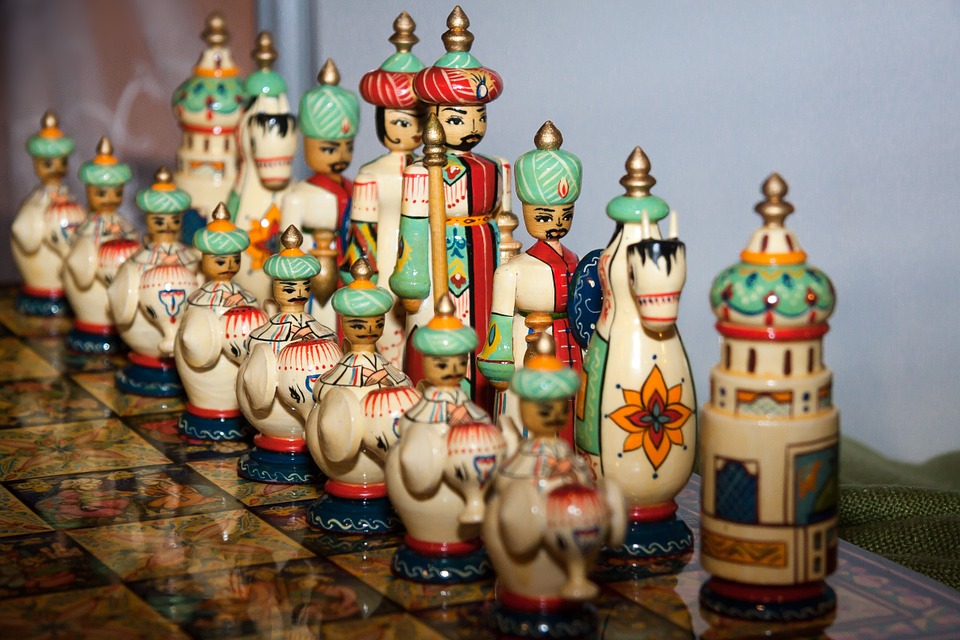} & \includegraphics[width=7cm]{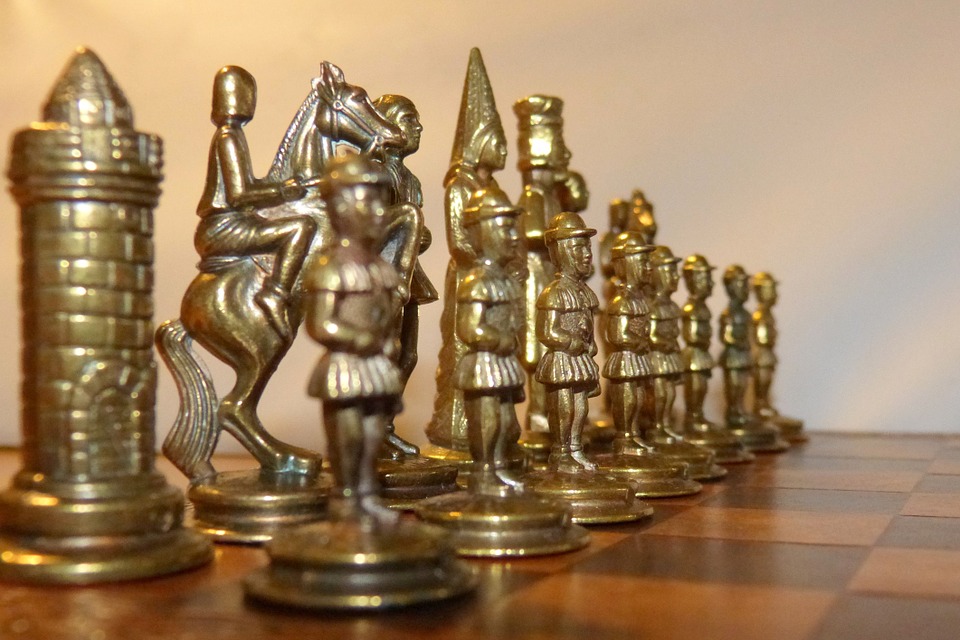}\\
	\end{tabular}
	\caption{Humans can intuitively recognize a number of different styles of previously unencountered chess pieces. This task is impossible for symbolic AI because coding explicit rules for this purpose is impossible. This example is in striking contrast with the game of chess rules of which can be summarised on a single A4.}
	\label{fig:chess}
\end{figure}

Unfortunately, not all problems lend themselves to being solved in this way. Moving a chess piece is a good example – I encourage you to stop reading for a moment and think about what rules you would have to apply to select and move a chess piece. You have to start from the beginning and recognize where the piece is currently located - a standard task for image recognition systems that are gaining so much traction these days.
What does a chess piece look like? Can you define rules that will allow the computer to recognize chess pieces from any angle under any light conditions? How about pieces with novel shapes that humans can instantaneously recognize? Consider the variety of chess pieces in figure \ref{fig:chess}. The difficulty of making these definitions is the reason why symbolic AI ultimately failed to translate to  ``easy’’ tasks such as image recognition despite the huge success in ``hard’’ tasks such as chess. At the same time the hard-coded rules of symbolic AI do not bear any resemblance to the fuzzy nature of biological intelligence. Machine Learning-based approaches bear more resemblance to human intelligence than those with symbolic AI, but is this really the driving force behind its huge success?

Machine Learning [ML], as the name suggests, is a branch of AI in which the machines learn to solve a problem themselves, as opposed to being given a set of explicit rules on how to solve the task. This is usually achieved by learning new rules through repeated attempts at the task, and feedback on the extent of success each time. This is the first fundamental similarity between the current progress in AI and biological Intelligence. We were not born with ability to recognize the objects surrounding us, neither (for the most part) were we told rules allowing us to identify those objects. Instead we have seen a very large number of objects that were sometimes labelled (think parents pointing out objects to a child) and learned from the experience.

Superficially, Machine Learning works in a very similar way. Computers are shown millions of pictures featuring various objects until they gradually learn what dogs, hats, cats etc. look like. Despite that this similarity may seem trivial and shallow it is by far the most important principle driving the current success of AI. While the similarity between artificial and biological intelligence goes much deeper, important differences remain. The machinery that has allowed computers to learn these powerful rules – Artificial Neural Networks [ANNs] - is directly inspired by neural networks in biological brains. In the rest of this article we will investigate how relevant this resemblance is to the abilities of this technology, and how deeply these similarities really go.

\section{Neurons}

\begin{figure}
	\includegraphics[width=\linewidth]{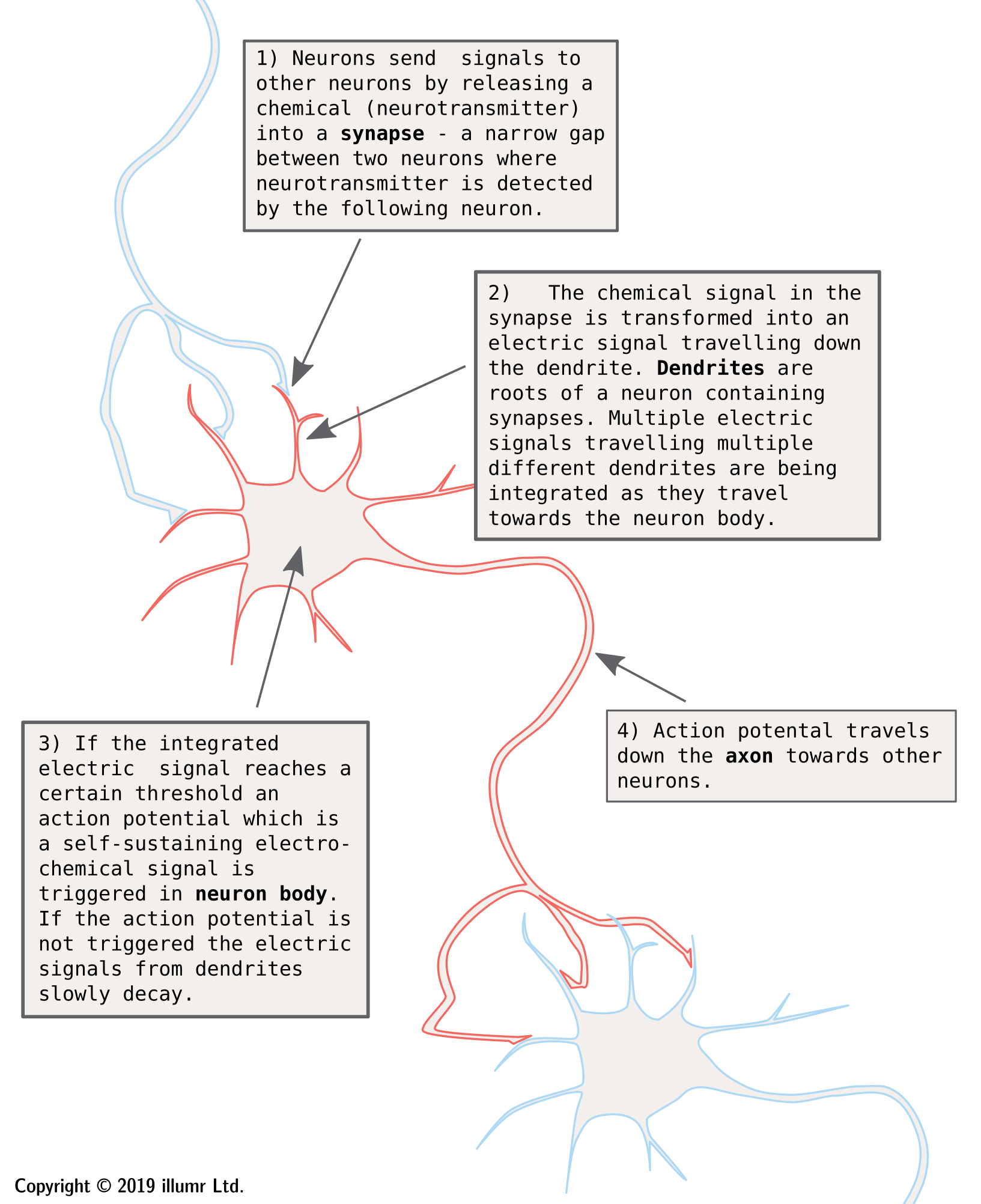}
	\caption{Basic structure of a Biological Neuron. A number of simplifications were made. For a proper description, please refer to a contemporary Neurobiology textbook.}
	\label{fig:biologicalNeuron}
\end{figure}

The tasks accomplished by ANNs resemble those accomplished by biological intelligence to some degree, and so do the building blocks of these two types of intelligent systems. (A basic description of a Biological Neuron is provided in figure \ref{fig:biologicalNeuron} for reference purposes.) The similarities  and differences between artificial and biological neurons have been used to argue for greater/lesser validity of methods in artificially replicating biological intelligence. However, this view is greatly simplified. In this section I will critically assess some of these claims one by one; however, the fundamental conclusion I will argue for is that the similarities between artificial and biological neurons are not a valid line of reasoning about the effectiveness of AI. 

In a nutshell, both types of neuron integrate the signal arriving from multiple other neurons, transform it with a non-linear function and output it to further neurons. 

Many people rightly point out that biological neurons operate using discrete signalling, while the artificial ones pass continuous numerical values.  Indeed, the action potential is binary - it is not known to have any graduation between on and off. However, individual action potential spikes do not usually significantly affect a biological neuron. Instead the spiking rate (frequency) seems to be important. Frequency, being a continuous value itself, doesn’t fundamentally differ from the information passed by artificial neurons. Therefore the validity of this difference is called into question.

The similarity of the type of non-linearity applied by both types of neurons is open to question too.  It is true that the binary activation function (step function) of biological neurons is poorly replicated 	by the sigmoidal non-linearities traditionally used in ANNs. However, if we consider the value output by artificial neurons to be akin to the firing frequency of biological ones, then we must instead compare the function which results from a period of firing. The superposition of many iterations of a fuzzy step function is well represented by a sigmoidal non-linearity, suggesting that there may actually be some degree of similarity in the activation function.

A large boost to the performance of ANNs was achieved not long ago by the introduction of a novel activation function. The rectified linear function, also called ReLU \cite{glorot2011deep}, helps neurons learn in situations that would formerly not be possible, due to its elegant mathematical properties.  The authors of this discovery also praised ReLU for its biological plausibility. The logic behind that argument was never clear to me.  While I do agree that the sigmoidal activation function doesn’t follow the activation function of biological neurons when taken at the face value, ReLU doesn’t even stay in a biologically plausible range. Instead it never saturates - meaning that more input always means more output without any limit. Biological neurons have a natural upper bound on how much they can fire - they get tired or even die when firing too much.

For a long time it has been thought that a neuron performs a single signal processing operation; however, more recently neuroscientists have revealed that the integration and simple logical processing of signal actually happens many times within a single neuron\footnote{The processing was found to happen already at the dendrite - before signals from many dendrites merge -  as opposed to being limited only to the cell body (cf. Figure \ref{fig:biologicalNeuron})} \cite{sardi2017new}. This finding makes the computation performed by a neuron several orders of magnitude more complex than it was thought to be just few years ago.  However, the computation is not different, it is just performed multiple times within one neuron. Using multiple artificial neurons where formerly a single one was used should therefore address this issue.

Neurons have been also found to be sensitive to temporal patterns in their stimulation \cite{fetz1997temporal} in a somewhat similar way to Morse code, which codes information as a binary signal in time. While artificial neurons do not have any such capability, any n-bit temporal pattern can be equally represented by a static pattern across n neurons. Therefore, replicating this finding only increases the quantity of the artificial neurons needed rather than their individual complexity.

While a significant number of arguments highlighting the differences between artificial and biological neurons do not stand up to scrutiny there are certainly many aspects of biological neurons that are not  replicated to any degree by artificial neurons. Biological neurons get tired or even die when firing for too long, their function is modulated by hormones, they recycle neurotransmitters, and they are affected by concentration of salt and potassium etc. The complexity is staggering and we are still far from understanding how biological neurons work. 

In fact, only a fraction of computation is performed by neurons. For decades, glial cells that make-up roughly three quarters of the brain were thought to be simply providing supporting function to neurons in the brain. However, they have been discovered to receive, process and distribute signals in a way only neurons were previously known to do \cite{haydon2001glia}. As of yet the implications of the computation performed by glial cells is very far from understood.

Furthermore, biologists like Dennis Bray have long argued that computation is not unique to brains, but rather omnipresent in living organisms, and that protein synthesis and other cellular processes can’t be denied the status of computation. However, these processes are limited to within a cell and cannot really affect the behaviour of an organism directly (despite being crucial to its functionality), therefore this should have little effect on our understanding of intelligence. This view is, however, being challenged by current research.

Last year a series of discoveries caused a stir in scientific community when fruit fly cells were found to synthesise virus-like proteins, and use them to communicate with other cells \cite{pastuzyn2018neuronal}. While it’s unclear whether similar phenomena exists in humans and the speed of such communication is likely extremely low, it is an example of computation within a cell other than a neuron being communicated across an organism. This is something we used to think was special about neurons, or later about the brain. 
In the light of the immense complexity of computation and communication in living organisms it seems futile to model the neuron to any degree of veracity hoping that it will provide us with technical advances. 

Artificial neurons will for a foreseeable future still be a very inaccurate models of biological neurons, which in turn are only a part of the machinery used to perform computation and communication necessary for biological intelligence. Does this mean that AI achieved by ANNs cannot in principle achieve similar capabilities to biological intelligence?

Structurally, the artificial  neuron is an extremely limited model for biological neuron. However, more detailed models of neuron (e.g. Numenta) or neural networks that communicate by discrete spiking (e.g. SpiNNaker), failed profoundly to deliver any advantage to AI. Structural similarity between neurons may therefore not be nearly as important property for AI as the function of a neuron.
Indeed as a basic unit performing elementary logical processing, the two types of neuron could be seen as very similar.

I have always understood the neural networks as universal function approximators whether they are biological or artificial (what else than an approximator could they be!). The precise structure of a function approximator cannot influence its computational capability as long as it is sufficiently general to approximate the desired function and possible to train in finite time. Given a sufficient number of neurons, ANNs can approximate any function \cite{hornik1989multilayer}. Therefore, there is no reason why they could not approximate a function performed by biological neurons. This is equally true about any other universal function approximator such as polynomials which do not aspire to any similarity with biological neurons. The reason why ANNs are gaining much more traction in AI than polynomials is that they are relatively easy to train, this however may not hold for long as novel research suggests that polynomials can beat ANNs on a number of ML benchmark problems \cite{specht1991general}. The change that brought about the advancement of polynomials was not the change of their structure or computational capability, but novel methods of training them to perform the desired function. In summary, it is in theory possible for ANNs to approximate biological functionality without similarity in the basic building blocks. The remaining question is whether we can train ANNs to perform the function of our interest. 

\section{Networks and Learning}

So far we have discussed only rudimentary properties of individual neurons. However, the way neurons are put together into networks and trained has historically had a larger impact on their performance than the structure of a single unit.

The first practical ANN was introduced by Frank Rosenblatt in the 1950s, called a Perceptron. It did not include any hidden layers and as such was profoundly limited. However, its parallel nature allowed it to easily learn simple logic - performing logic on parallel inputs is much easier than on serial, as no memory is required to store intermediate values. Deep layers were not around yet, mostly because no one knew how to train them. Minsky and Pappert demonstrated in 1969 that this basic kind of ANN architecture can never solve linearly inseparable problems (e.g. perform logical operations like exclusive OR) and concluded that they will never be able to solve any actually interesting problem. In their book they wrongly conjured that this problem is common to all ANNs, which caused a huge decrease of interest/funding in ANNs, essentially halting the research in the area for almost two decades.

It took until the 1980s to salvage what was left of ANNs. An international group of researchers (David Rumelhart, Jeff Hinton and others) came up with a method to train deep ANNs called backpropagation. This immediately disproved the conjecture of Minsky and Pappert and kicked-off a large research effort into ANNs. It is worthwhile to note that backpropagation is perhaps the aspect of ANNs most widely criticised for being biologically implausible.	

Backpropagation relies on propagating errors backwards through a deep ANN to correct/train the deep layers. The algorithm of backpropagation roots in automatic differentiation, a method developed by Newton at a time when there was essentially no understanding of the biology underlying intelligence. Claiming biological inspiration as its inception would therefore be silly.

The main criticism of the biological plausibility of backpropagation is focused on the requirement to feed signals backwards, which biological neurons are known to be unable to do\footnote{This is not entirely true as some marginal cases of backwards signalling have been reported.}. However, neuroscientists have known for a long time that neurons are not a uniform mass of homogeneous material, but instead they repeatedly appear in structured assemblies that have feedback connections \cite{bastos2012canonical}. Therefore it is entirely plausible that at least in some brain areas neurons exist in pairs where one feeds forwards while the other one is feeding back making backpropagation entirely possible within a neural assembly. This being said, there has been no convincing demonstration of backpropagation of error in brain so far - something we should expact to be relatively easy to observe if it is a prominent feature of biological neural networks.

Backpropagation was a necessary advancement for deep neural networks, as the depth is one of the most powerful concepts brought to ANNs. Depth allows for hierarchical representation of a problem - solving smaller problems (e.g. detection of edges) first before moving onto bigger ones (e.g. recognizing an object from a collection of edges). Hierarchical nature is also a prominent feature of information processing in brains \cite{riesenhuber1999hierarchical}, providing an elegant alignment between biological and artificial intelligence.

Another large improvement to the capabilities of ANNs was the introduction of convolution as one of the main components \cite{lecun1995convolutional}. Convolution essentially means moving a filter across the data to identify features. This brought a large advantage to the training of ANNs as it hugely decreases the number of free parameters to be tweaked. Interestingly, it is well evidenced that similar principles are being applied in the brain \cite{glezer1973investigation}, and is perhaps are one of the crucial phenomena allowing brain to process visual stimuli very efficiently.

So far we have seen that the criticism of backpropagation for it’s biological implausibility is shaky (while no evidence for backpropagation within biological neural networks has been found so far), and that convolution seems biologically well founded. However, the last aspect of contemporary ANNs does not follow this line of reasoning. Initialization of weights by an identity matrix \cite{le2015simple} is a technique that facilitates the training of deeper networks. It works because initially the deep layers just pass on their inputs unaltered – a better starting point than learning from a random complex transformation. As you might expect this doesn’t lend itself to any meaningful comparison with the brain, illustrating that biology, while being a useful inspiration to ANN architecture is by far not the only means of advancement in AI.

In conclusion, the main advances driving the current success of ANNs are focused on how ANNs are trained, not what exactly they are. These advances are a mix of mainly pragmatic engineering choices and some questionable biological inspiration which is also a good engineering choice.

\section{Intelligence}
Comparison of the machinery of ANNs with biological neural networks has so far escaped simple conclusions. Here I will try to compare the intelligence (high-level functionality) that arises from artificial and biological neural networks. To obtain truly conclusive evidence of a deeper relationship between ANNs and biological intelligence it would be useful to see biological-like behaviour emerge from a system that we have not explicitly designed to do so. Using emergent properties of simple neural networks to explain little understood properties of the brain, such as the idiosyncratic functional divisions between brain areas, was the agenda with which I first entered academic research myself.

An early motivation for research into ANNs, whilst they were not yet powerful enough to address practical tasks, was the similarity in their performance to some aspects of human behaviour. David Rumelhart and Jay McClelland started an entire research program within the area of psychology by exploiting these similarities. They demonstrated that when reading partially obscured words, ANNs and humans infer  the obscured parts in a strikingly similar fashion \cite{mcclelland1981interactive}. Later on, Matt Lambon-Ralph demonstrated that when exposed to a new language, ANNs pick-up words in a similar way to human children \cite{ellis2000age}. There are many other examples demonstrating similarities between very simple ANNs and human behaviour. 

The similarity between the performance of simple ANNs and humans is not very impressive given that these particular networks cannot manage to solve anything beyond toy tasks, and are therefore very far from modelling anything like human intelligence. Recently, however, ANNs gained the ability to perform some basic real-world tasks. Research on these demonstrated  similarities in the internal state of a Deep Convolutional Neural Network and the human brain when processing the same visual stimuli \cite{gucclu2015deep}. Specifically, it has been demonstrated that the representations in both systems are based on hierarchical pattern recognition.

These results should, however, surprise no one. It has been known to neuroscientists for a long time that the visual processing in the brain is akin to hierarchical pattern recognition. The ANNs of Guclu and Gerven are only a mathematical specification of a  hierarchical pattern recognition system. The similarity therefore reflects a design decision, not an emergent property, and can be hardly used to argue deeper similarities between ANNs and biological intelligence. In fact, most of the emergent properties make a convincing argument against functional similarities between ANNs and biological intelligence.

It has been long argued that the amount of data required to train a contemporary ANN is in a clear juxtaposition to animals being able to learn from very few observations. In 2017 Google Deep Mind published a paper in Nature boasting that they managed to train a network called AlphaGoZero within only 36 hours, that not only beats any human Go player, but also AlphaGo – the previous strongest Go player \cite{silver2017mastering}. A deeper dive into the aforementioned Nature paper, however, reveals that in those 36 hours AlphaGoZero played 4.9 million games. This roughly corresponds to 559 years of non-stop Go at human speed, further emphasising how distant are the data requirements of ANNs are from humans learning the same task.

A similar issue was well-documented by Dubey and colleagues in 2018 \cite{dubey2018investigating}. ANNs were found to need much prolonged training to learn the basics of playing Montezuma’s Revenge (a vintage video game) compared to people. After mixing-up the textures in the game (so that, for example, the sky had a ground texture) the game stopped making sense for humans while the ANNs learned the altered game in approximately same amount of time as the original. 
This finding is suggestive of a significant difference between the way ANNs and humans tackle novel situations. Humans apply their past knowledge to novel situations – for example they may attempt to climb a ladder rendered in a game, despite the immense visual difference from a real-world ladder. However, this makes it very difficult for humans to make sense of a sky that looks like a ground. The ANNs took longer to learn the game in the first place because they are unable to apply knowledge obtained in different contexts (e.g. real-life), and they were also uninhibited by prior assumptions when learning something new. It is far from clear how to make ANNs apply knowledge gathered in different contexts\footnote{There has been some progress in the field of Transfer Learning which aims at enabling ANNs to perform these tasks, however even the most advanced systems are profoundly limited to specific tasks/contexts.}. 

This problem – lack of ability to perform transitive inference - is in my view the largest barrier for advancement of AI. Hector Levesque succinctly summarises the problem in his book titled ``Common Sense, the Turing Test, and the Quest for Real AI’’ by asking a question: How would crocodile do in a steeplechase? While finding an answer may seem trivial to you, it requires  the amount of transitive inference as of yet never seen in any ANN, bar those specifically crafted for this purpose - which are unable to perform other functions to a reasonable standard.

Even if ANNs could perform transitive inference, their prior knowledge would still be fundamentally different from ours, probably leading to profoundly different behaviour. This is rooted in the fact that the training sets utilized by ANNs are so different to those available to animals. Linda B. Smith has been spearheading a collection of headcam footage from toddlers \cite{bambach2018toddler}, which is as far from ML object recognition datasets as it could be. The most prevalent objects are faces, and somewhat abstract imagery such as wall corners or branches pictured against the sky; a stark difference to the general cross-section of images from the internet used to train ANNs.  I see a significant potential in ANNs trained on this footage to perform in a more similar way to humans than any previous models.

\begin{figure}
	\centering
	\begin{tabular}	{lll}
		
		\includegraphics[height=3cm]{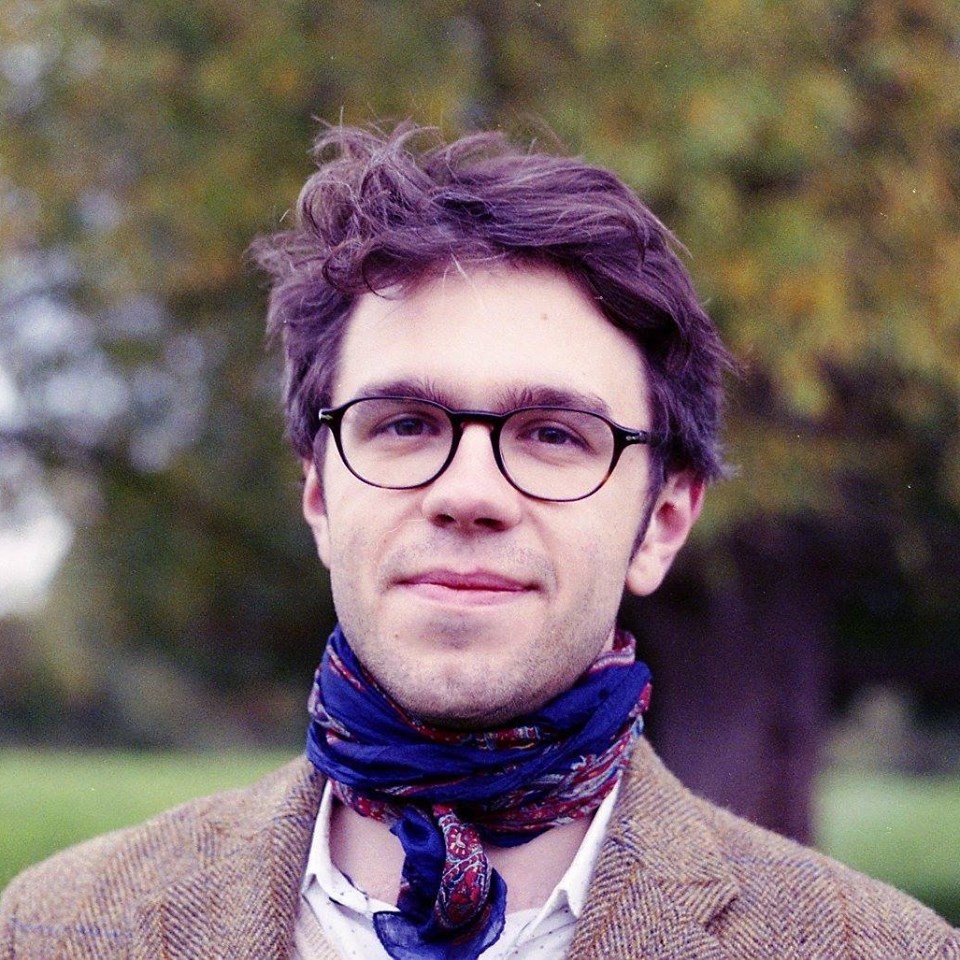}&	&	\includegraphics[height=3cm]{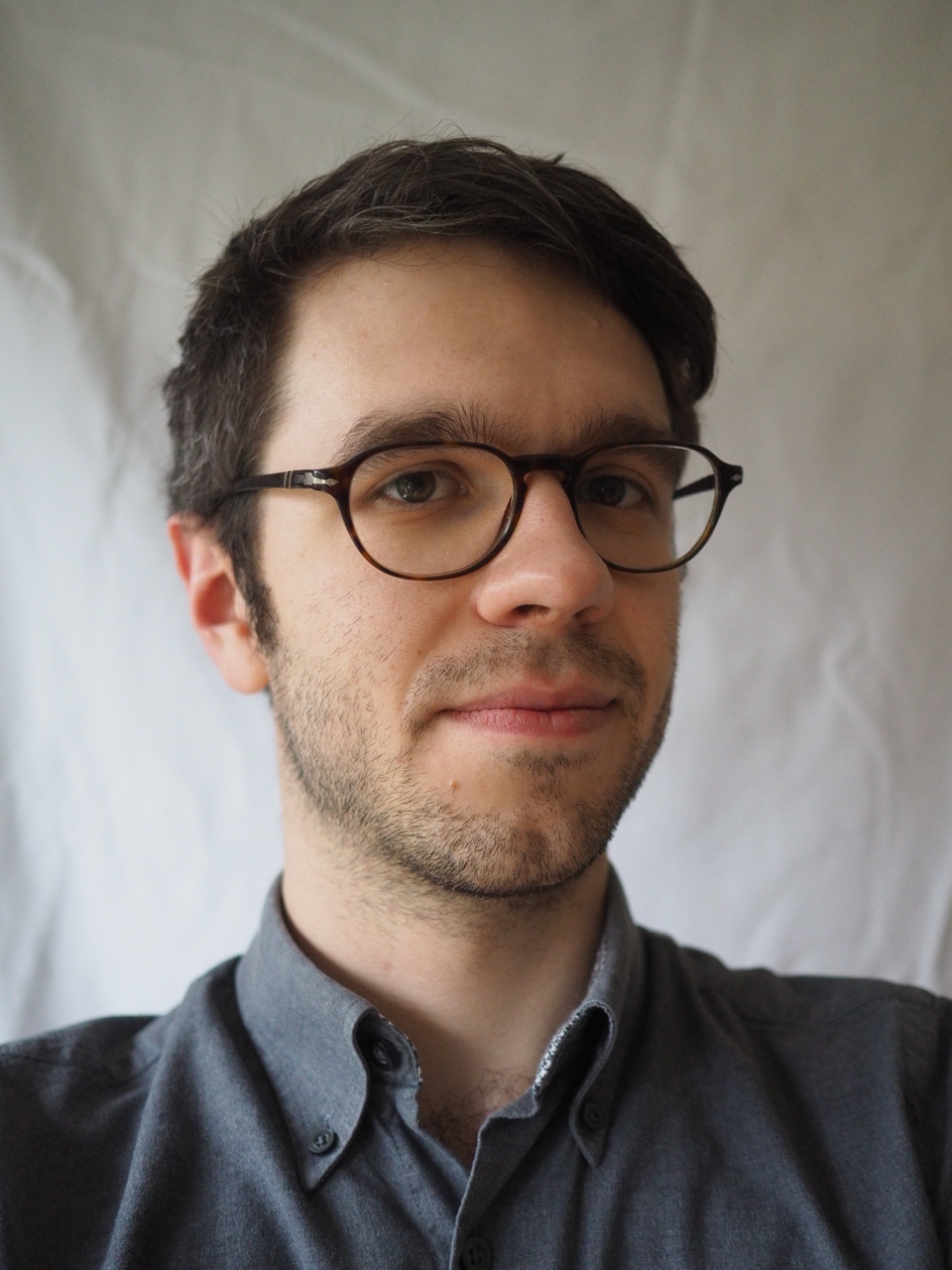}\\
		99\% man & &84\% woman\\
		51\% age 27 && 61\% age 23 \\
		98\% white & &40\% hispanic, 18\% black\\
	\end{tabular}
	\caption{Two attempts to judge author's demographics made by a company offering demographics-estimation by ANNs (Clarifai) from a photograph. The shocking difference in the estimated demographics suggests that the features used are fundamentally different from those used by humans for the same task.}
	\label{fig:clarifai}
\end{figure}

Transitive inference might simply be a functionality that we haven’t yet achieved. However, there are good reasons to believe that even the current success of ANNs in particular tasks is achieved in a way very different to humans. Consider the two attempts (shown in figure \ref{fig:clarifai}) I made to estimate my own demographics from an image, by a demographics-estimating product offered by the company Clarifai. The first attempt was fairly accurate, concluding the picture 99\% likely  to be a man, 51\% likely to be age 27 and 98\% likely to be of white ``multicultural appearance''.  The second attempt resulted in  84\% likely  to be a woman, 61\% likely to be age 23 (thanks Clarifai!) and 40\% likely to be of hispanic ``multicultural appearance''. I was also judged to be 18\% likely to be black or African American.

The argument that I’d like to make here is not that Clarifai's demographics-estimating product is bad (I leave making that judgement to the reader), but rather that the principles on which it estimates the demographics must be fundamentally different to how humans accomplish the task. Sure, people have varying ability to estimate the demographics; however, if someone estimates the demographics from the first picture as well as Clarifai did, it seems highly unlikely they would fail so miserably on the second picture.

The stark difference in features used for visual object recognition between ANNs and humans is perhaps best illustrated by adversarial examples \cite{goodfellow2014generative}. These examples are tiny perturbations made to an original image that trick an ANN into identifying an object as something entirely different to what it did before the perturbation. Adversarial examples won’t trick humans as we can mostly not even detect the perturbation. It turns out the same is true the other way around; ANNs are not tricked by human visual illusions \cite{williams2018optical}.

The intelligence displayed by ANNs can be very similar to that of humans as long as the ANN was successfully trained for that purpose. For instance an ANN trained to identify images will be good at that task, just like biological intelligence is, but it will hardly replicate the way how we accomplish that task as that was not the objective of the training. Therefore the performance on a task the ANN was trained for can never suggest similarity in the basis of intelligence, but rather success in the engineering task. Furthermore, ANNs require orders of magnitude more data to learn from than humans, and there is no known way how to make them successfully perform transitive inference yet – a task we perform nearly all the time to our great advantage.

\section{Conclusion}

This article has compared artificial and biological Intelligence from a number of angles. 
I proposed a view that there is no reason to expect different computational properties from two different universal function approximators – which both artificial and biological neurons undoubtedly are. Therefore, there is a decent reason not to worry about the impact of huge structural differences between biological and artificial neurons on the intelligent behaviour that emerges from this machinery. What seems more important is how learning of the desired function is performed. There is very little known resemblance between artificial and biological intelligence in this respect, except that both are capable of succesfully learning many of the tasks we are interested in. The lack of resemblance between these two types of intelligence is highlighted by the huge differences in performance on any task the ANNs were explicitly trained to mimick biological intelligence.

A significant proportion of the widely circulated criticism relating to lack of biological plausibility of the ANN machinery seems to stem from a lack of up-to-date knowledge of neurobiology. However, the difference between the two types of neurons remains very significant even after doing away with broken arguments, especially when we consider the perplexing complexity revealed by contemporary neurobiology.
The difference is exacerbated by the fact that, contrary to how they are presented in popular knowledge, neurons are not the only cells capable of computation and communication. Such capabilities likely spread beyond the nervous system or brain, and are mostly unknown. Furthermore, currently even the properties of neurons are poorly understood by science. Focusing on the veracity of the representation of biological neurons therefore seems rather naive.

The story gets little more complicated when we look at the structure of ANNs beyond individual neurons. Some of the technological advances that are driving the current ML revolution are clearly biologically implausible (identity matrix weight initialisation etc.), and at the same time are crucial in allowing ANNs to achieve their current performance. On the other hand, implementation of some ideas that are biologically well-grounded, such as hierarchical pattern recognition, have made a crucial and positive contribution to the performance of ANNs.

In the last section I discussed the functional differences between intelligence of ANNs and animals. There has been a long-standing research effort demonstrating similar patterns of performance between trivial ANNs and humans. However, it is unclear what the performance of toy models on toy tasks can tell us about true intelligence. 

Recently efforts have been made to identify common patterns between ANNs that can actually solve real-world tasks, and the human brain.  In my view these findings have so far failed to identify any patterns of similarity that are not simple design decisions made during construction of the ANN. To the contrary, there is a plenty of evidence suggesting the two types of intelligence are fundamentally different. The amount of data needed for learning differs by several orders of magnitude, and there is no capability to use previous knowledge or perform transitive reasoning. Finally, the features exploited to achieve a task are fundamentally different.

I would like to end this article on a less damning note. Despite the profound differences in both structure and function between ANNs and biological intelligence it is necessary to remember that some of the fundamental principles (parallel, hierarchical, non-linear etc.) of biological intelligence applied in the ANNs led to their astonishing success, and there might be more to follow.

\bibliographystyle{plain}
\bibliography{bibliography}

\end{document}